\documentclass{article}

\usepackage{microtype}
\usepackage{graphicx}
\usepackage{subfigure}
\usepackage{booktabs} 

\usepackage{hyperref}

\usepackage[accepted]{icml2024}

\usepackage{amsmath}
\usepackage{amssymb}
\usepackage{mathtools}
\usepackage{amsthm}
\usepackage{multirow}

\usepackage[capitalize,noabbrev]{cleveref}

\theoremstyle{plain}

\theoremstyle{definition}

\theoremstyle{remark}

\usepackage[textsize=tiny]{todonotes}

\icmltitlerunning{Test-Time Degradation Adaptation for Open-Set Image Restoration}

\begin{document}

\twocolumn[
\icmltitle{Test-Time Degradation Adaptation for Open-Set Image Restoration}




\begin{icmlauthorlist}

\icmlauthor{Yuanbiao Gou}{univ}
\icmlauthor{Haiyu Zhao}{univ}
\icmlauthor{Boyun Li}{univ}
\icmlauthor{Xinyan Xiao}{comp}
\icmlauthor{Xi Peng}{univ}

\end{icmlauthorlist}

\icmlaffiliation{comp}{Baidu Inc., Beijing, China}
\icmlaffiliation{univ}{
College of Computer Science, Sichuan University, Chengdu, China}

\icmlcorrespondingauthor{Xi Peng}{pengx.gm@gmail.com}

\icmlkeywords{Image Restoration, Test-Time Adaptation, Diffusion Model}

\vskip 0.3in
]



\printAffiliationsAndNotice{}  

\begin{abstract}
In contrast to close-set scenarios that restore images from a predefined set of degradations, open-set image restoration aims to handle the unknown degradations that were unforeseen during the pretraining phase, which is less-touched as far as we know. This work study this challenging problem and reveal its essence as unidentified distribution shifts between the test and training data. Recently, test-time adaptation has emerged as a fundamental method to address this inherent disparities. Inspired by it, we propose a test-time degradation adaptation framework for open-set image restoration, which consists of three components, \textit{i.e.}, i) a pre-trained and degradation-agnostic diffusion model for generating clean images, ii) a test-time degradation adapter adapts the unknown degradations based on the input image during the testing phase, and iii) the adapter-guided image restoration guides the model through the adapter to produce the corresponding clean image. Through experiments on multiple degradations, we show that our method achieves comparable even better performance than those task-specific methods. The code is available at https://github.com/XLearning-SCU/2024-ICML-TAO.
\end{abstract}

\section{Introduction}
\label{intro}

In recent years, significant advances have been made in the realm of image restoration, and have demonstrated remarkable capabilities in addressing a multitude of image degradations~\cite{gdp, jiang2023autodir, msanet, code, yoly}. However, a common limitation among most existing methods is their reliance on assumptions rooted in a close-set scenario, \textit{i.e.}, the test degradations closely resemble those encountered during the pretraining phase. This assumption restricts the applicability of these methods to specific types of degradation, hindering their adaptability to a broader range of real-world scenarios where diverse and unforeseen degradations may arise.

\begin{figure}
    \centering
    \includegraphics[width=0.99\linewidth]{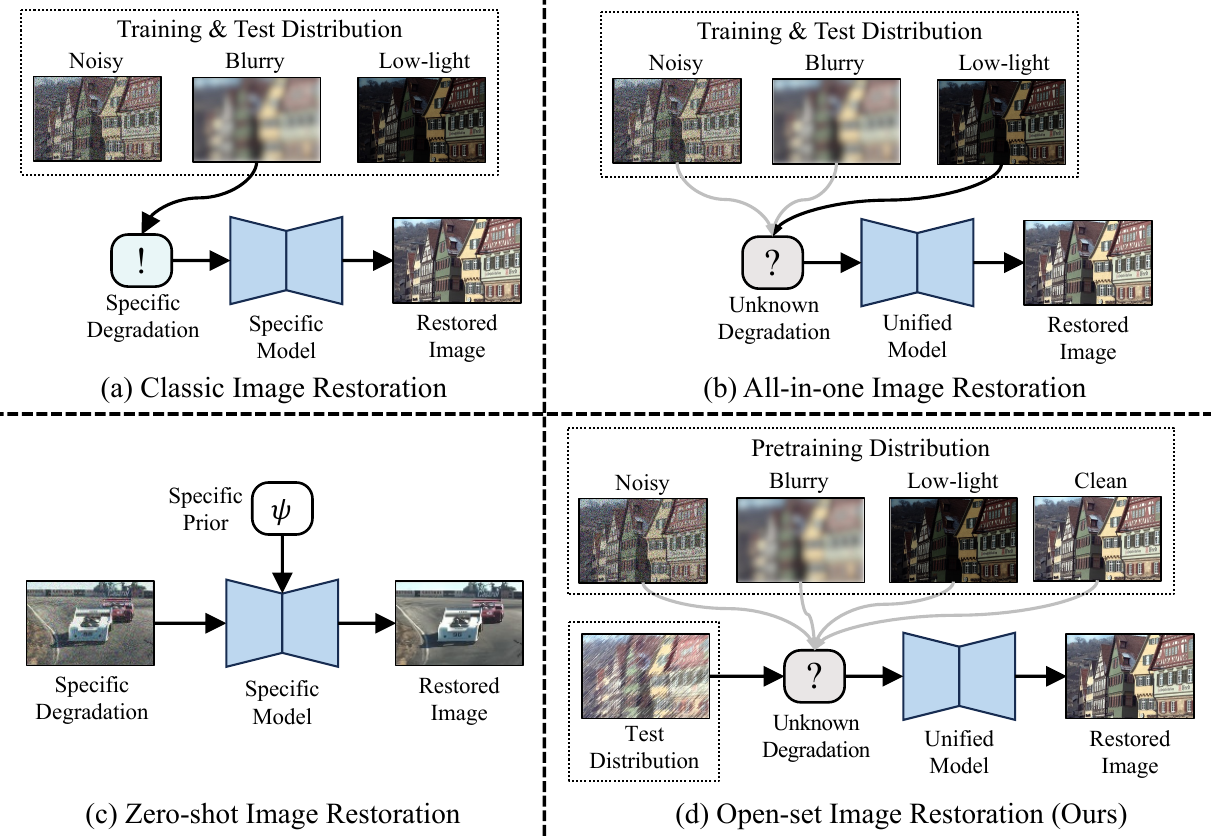}
    \vspace{-0.2cm}
    \caption{The differences of image restoration (IR) tasks. To be specific, (a) Classic IR works in a close-set scenario where the training and test degradations are the same and known, and customizes a specialized model for each one. (b) All-in-one IR also works in a close-set scenario where the training and test degradations are the same but unknown, and addresses them through a unified model. (c) Zero-shot IR focuses on recovery from single degraded image, which is free from the training degradations, but often requires priors about the test degradations in advance. In contrast, (d) OIR works in an open-set scenario where the test degradations are unknown and different from the pretraining ones. This is analogous to the challenge in natural language processing, which applies the pre-trained large language model to the various downstream tasks not predefined during the pretraining phase.}
    \label{fig:setting}
\end{figure}

To break the above assumption, this work delves into the more challenging and less-touched problem of open-set image restoration (OIR). In contrast to conventional close-set approaches, where models only address the specific degradations encountered during the pretraining phase, OIR poses a formidable challenge by requiring models to handle the unknown degradations absent from the training data. This shift in focus aims to push the boundaries of image restoration, fostering the development of models that exhibit adaptability and resilience in the face of diverse and unexpected degradations. By tackling the OIR problem, this work seeks to pave the way for more robust and versatile solutions applicable to complex degradations in the real-world scenarios.

The essence of OIR lies in addressing the challenges posed by real-world scenarios, where the test data may differ significantly from the training data. This could involve variations in degradation types and levels, lighting conditions and so on that were not explicitly covered in the training data. To tackle this challenge, test-time adaptation (TTA) has recently emerged as an effective methodology to address the inherent disparities between the test and training data~\cite{tent}. Specifically, it adapts the pre-trained model during the testing phase based on the specific characteristics of the test data, allowing the model to perform better on a wider range of input scenarios. This is particularly useful in OIR where the test data may vary from the training data, and the pre-trained model should to adapt to unknown and unseen degradations for optimal performance.

Building on this inspiration, we present a \textbf{T}est-time degradation \textbf{A}daptation framework for \textbf{O}pen-set image restoration, dubbed TAO, which harnesses the idea of TTA to provide a robust solution for OIR. Specifically, TAO consists of three components, \textit{i.e.}, a pre-trained image diffusion model (PDM), a test-time degradation adapter (TDA), and the adapter-guided image restoration (AIR). In this framework, PDM is adopted as the foundation model for OIR due to the following considerations. First, PDM captures rich knowledges of generating various high-quality visual scenarios, which could be regarded as a generic pretraining for OIR targeting at producing clean images. Second, PDM is degradation-agnostic and any degradations in the test data could be considered as unforeseen. Note that the test phase of PDM is referred to the process of image generation, \textit{i.e.}, the step-by-step denoising process.

After each denoising step, TDA and AIR are sequentially conducted for adapting to open-set scenarios and guiding image restoration, respectively. Specifically, TDA employs a learnable adapter during the test phase for adapting PDM on the test degraded image. This adapter is devised for domain alignment that aligns the generative domain of PDM to the degraded domain of the test image. In this way, the generated clean image could be translated to the corresponding degraded one, which could be further supervised towards the test degraded image, and in turn updating the generated clean image. AIR is elaborated to conduct this supervised updating for image restoration. In brief, AIR is inspired by the observation that PDM exhibits an intriguing temporal dynamic~\cite{choi2022perception} during the step-by-step denoising process. Therefore, AIR dynamically and accordingly adjusts the supervision strategies at different denoising steps for achieving a better performance.

Overall, our contributions could be summarized as follows.
\begin{itemize}
    \item As far as we know, this could be one of the first work that explicitly studies OIR and discovers its essence, \textit{i.e.}, the unidentified distribution shifts between the test and training data.
    \item This work reveals that TTA is an effective methodology for OIR, by adapting the pre-trained model on the test data during the test phase, to address the inherent disparities between the test and training data.
    \item This work presents a test-time degradation adaptation framework for OIR, where two components are devised to adapt open-set scenarios and guide image restoration. Experiments show the effectiveness of this framework.
\end{itemize}

\section{Related Work}
\label{sec:rel_work}

In this section, we briefly review recent advances in related topics, which mainly involves test-time adaptation, all-in-one image restoration, and zero-shot image restoration.

\textbf{Test-time adaptation} has shown to be effective at tackling distribution shifts between test and training data, by adapting the pre-trained model on test samples. In the past several years, it has gained increasing attentions and plentiful methods have been proposed. For example, \cite{tent, yang2024renata} updated the specific model parameters through the test samples by resorting to the unsupervised objectives. Similarly, \cite{wang2022continual, gan2023decorate} aimed to solve the continually changing distribution shifts along the test time. Moreover, \cite{niu2023towards, zhou2023ods} considered more challenging and practical adaptation settings such as single sample, label shifts, and mixed domain shifts. In addition, \cite{shin2022mm, lee2023tta} shifted their focuses on applications beyond image recognition, such as semantic segmentation and pose estimation.

This work focuses on application of solving OIR through TTA methodologies. In brief, we incorporate PDM with an adapter which is updated for adapting to the test degraded image during the test phase. Besides, we also consider the challenging adaptation setting of single test sample.

\textbf{All-in-one Image Restoration} is an emerging issue which has been attracted more and more attentions. Specifically, it aims to address the unknown degradations within a predefined set through one model. For example, \cite{air} firstly studied this problem and proposed to automatically extract the degradation representation from the unknowingly degraded images for assisting their restoration. \cite{promptir} aimed to encode the degradation information into prompts which are then used to dynamically guide the restoration. \cite{jiang2023autodir} focused on using the generated text prompts to guide the latent diffusion model for restoring images with unknown degradations. Although remarkable progress has been achieved, they usually need to be trained on a predefined set of degradations, and recover images from the degradations that are unknown but within this set. In other words, they could only be applied to the close-set scenarios.

In contrast, this work is devoted to address OIR challenge, wherein the test degradations are unknown during the test phase, meanwhile unforeseen during the pretraining phase. Note that OIR is established on the pre-trained model for image restoration, and adapts the model to the downstream tasks (\textit{i.e.}, unknown and unforeseen degradations) during the test phase. This work empirically regards the diffusion model~\cite{ddpm} as a generic pretraining task for various image restoration tasks.

\textbf{Zero-shot Image Restoration} focuses on recovering the clean image from a single degraded one, without relying on paired clean-degraded images for training. Classic methods are usually devised for specific tasks. For example, \cite{shocher2018zero} performed super resolution by downsampling the low-resolution image to construct the lower-low training pairs. \cite{li2021you} trained dehazing networks by decoupling and coupling the haze image through the physical model of haze formation. In recent, some works focused on methods with more general purpose. For instance, \cite{ddrm, ddnm} applied the decomposition approaches on the predefined linear degradations, and used the diffusion model to address multiple tasks of image restoration. Similarly, \cite{gdp} introduced a conditional guidance for eliminating linear and blind degradations through the predefined and optimizable linear degradation operators, respectively.

This work is different from the above methods on both the problem and solution. On the problem, zero-shot focuses on the restoration from single degraded sample, while open-set focuses on dealing with the unknown degradations that were unforeseen during the pretraining phase, where the degraded sample does not have to be a single one. On the solution, the above methods usually exploit additional degradation priors to assist restoration from single degraded image, while our method only introduces a degradation-agnostic adapter for all unknown and unforeseen degradations.

\section{The Proposed Method}
\label{sec:method}

In this section, we first briefly review the preliminary knowledges, and then progressively illustrate the proposed TAO and its two core components of TDA and AIR.

\begin{figure*}[!htbp]
    \centering
    \includegraphics[width=0.99\textwidth]{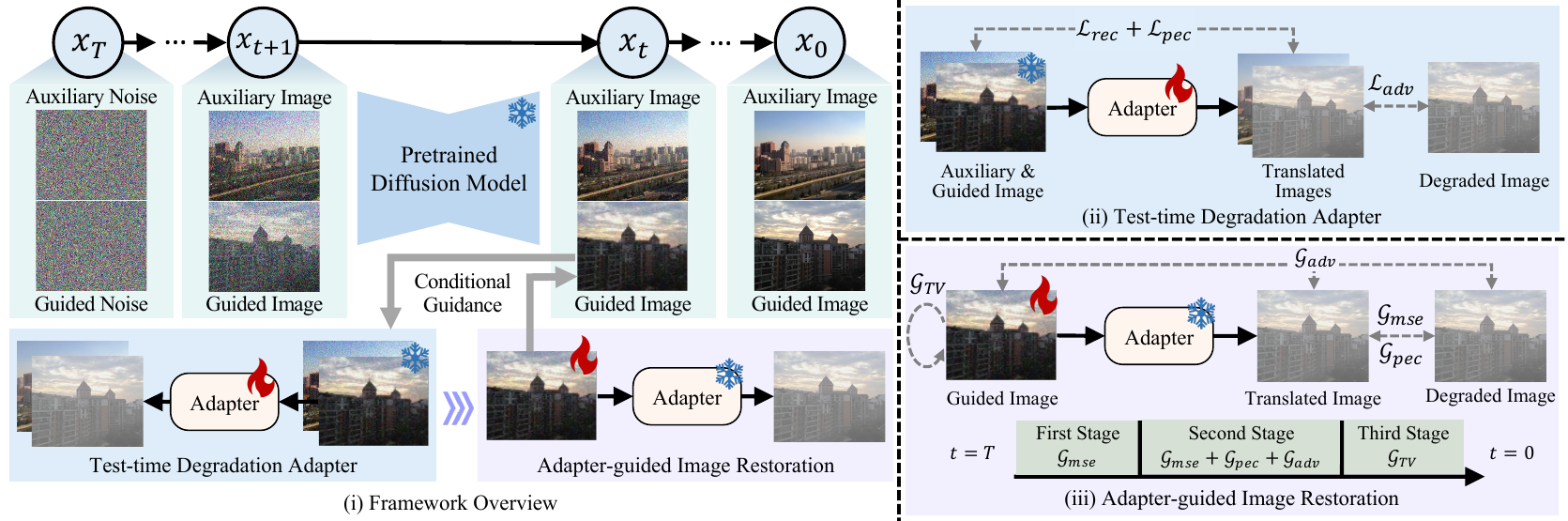}
    \vspace{-0.2cm}
    \caption{Overview of the proposed method, which exploits (i) a PDM as the generic pre-trained model for OIR. After each denoising step, it first performs (ii) TDA for adapting to the unknown and unseen degradations posed by open-set scenarios, and then conducts (iii) AIR for optimizing the guided image towards the restored clean image. Note: the snowflake icon indicates the image or model is fixed, and the flame icon indicates the image or model will be updated through the gradients.}
    \label{fig:arch}
\end{figure*}

\subsection{Preliminary}

\textbf{Open-set image restoration} is devoted to handle the unknown degradations that were also unforeseen during the pretraining phase. Mathematically, for a given model $M$ pretrained on the distribution $D_{train}$, OIR requires to restore a test degraded image $y \sim D_{test}$ so that $L(M(y), x)$ is minimal, wherein $D_{test}$ is unknown and different from $D_{train}$, $L$ is a function to measure the discrepancy, and $x$ is the clean version of $y$. Formally, OIR essentially poses a problem of the unidentified distribution shifts between the test and training data. For OIR, two fundamental principles should be followed. (i) Unknown: OIR is agnostic to degradations during the test phase, and any prior knowledge about specific degradations is prohibitive. ii) Unforeseen: OIR is independent of a specific pretraining task, and only requires the test degradations were not presented for pretraining.

\textbf{Test-time adaptation} aims to improve the model's performance on new, unseen data by adjusting its parameters based on the test samples. Mathematically, for a given model $M$ pretrained on the distribution $D_{train}$, TTA optimizes some parameters $\theta$ from $M$ on the test samples $y \sim D_{test}$ through the objective function $\min_{\theta} L(y;\theta)$, wherein $D_{test}$ is different from $D_{train}$. For TTA, two fundamental principles should be followed. (i) TTA starts with a pre-trained model which has learned general knowledges for the target task. (ii) TTA works on the test data where the training pairs are unavailable. In this work, we adopted the diffusion model as the pre-trained model, the generated and degraded images as the training pairs to implement TTA for OIR task.

\textbf{The Diffusion Model.} Given a data distribution $x_0 \sim q(x_0)$ and a noise distribution $x_T \sim \mathcal{N}(0,I)$, diffusion model~\cite{ddpm} defines a $T$-timestep diffusion process, which corrupts $x_0$ to $x_T$ by sequentially adding the random noise $\epsilon \sim \mathcal{N}(0,I)$, and a $T$-timestep denoising process, which recovers $x_T$ to $x_0$ by progressively eliminating the noise $\epsilon$. Importantly, there is an elegant property in the diffusion process, \textit{i.e.},
\begin{equation}
    x_t = \sqrt{\Bar{\alpha}_t} x_0 + \sqrt{1 - \Bar{\alpha}_t} \epsilon,
    \label{eq:deg}
\end{equation}
where $t \in \{0, ..., T\}$, $\Bar{\alpha}_t=\prod^t_{i=0}\alpha_i$, $\alpha_i = 1-\beta_i$ and $\beta_i$ is the variance of the $i$th timestep. Based on this formula, the diffusion model learns a denoiser for the denoising process via the following objective, \textit{i.e.},
\begin{equation}
    \triangledown_{\theta} \parallel \epsilon - \mathcal{Z}_{\theta}(\sqrt{\Bar{\alpha}_t} x_0 + \sqrt{1 - \Bar{\alpha}_t} \epsilon, t) \parallel,
\end{equation}
where $\mathcal{Z}_{\theta}$ is the denoising network, and $t$ is randomly sampled during the training phase. For a well-trained diffusion model, denoising process progressively yields $x_{t-1}$ from $x_t$ in terms of $\epsilon_t=\mathcal{Z}_{\theta}(x_t, t)$ until $t=0$. According to Bayes Theorem, the $x_{t-1}$ could be sampled through the following process, \textit{i.e.},
\begin{equation}
\begin{aligned}
    q(x_{t-1} | x_t) &= \mathcal{N}(x_{t-1};\mu(x_t, \hat{x}_0), \tilde\beta_t), \\
    \mu(x_t, \hat{x}_0 ) &= \frac{\sqrt{\Bar{\alpha}_{t-1}}\beta_t \hat{x}_0 + \sqrt{\alpha_t}(1-\Bar{\alpha}_{t-1}) x_t}{1 - \Bar{\alpha}_t}, \\
    \hat{x}_0 &= \frac{x_t - \sqrt{1 - \Bar{\alpha}_t} \epsilon_t}{\sqrt{\Bar{\alpha}_t}}, 
    \; \tilde\beta_t = \frac{1-\Bar{\alpha}_{t-1}}{1-\Bar{\alpha}_t}\beta_t,
\end{aligned}
\end{equation}
where $\hat{x}_0$ is the estimated output from $x_t$ in terms of Eq.(\ref{eq:deg}). To guide the $x_{t-1}$ towards the desired output $x_0$, the conditional guidance $y$ could be introduced through the following formula~\cite{guidiff, gdp}, \textit{i.e.},
\begin{equation}
    q(x_{t-1} | x_t, y) \propto N(x_{t-1}; \mu(x_t, \hat{x}_0) + s\mathcal{G}, \tilde\beta_t),
    \label{eq:gui}
\end{equation}
where $s$ is the scale of guidance, $\mathcal{G}=\triangledown_{\hat{x}_0}\log p_{\phi}(y|\hat{x}_0)$, and $p_{\phi}(\cdot)$ is a model bridging the gaps between $\hat{x}_0$ and $y$. For image restoration, the test degraded image is introduced as the $y$, and the $x_0$ is the desired clean output.

\subsection{Framework Overview}

To achieve OIR through PDM, the key lies in implementing $p_{\phi}(\cdot)$ to produce the appropriate guidances of gradients $\mathcal{G}$, guiding the generated image $\hat{x}_0$ towards the clean image $x_0$ of the test degraded image $y$. In this work, we implement $p_{\phi}(\cdot)$ through TDA and AIR, which are devised for open-set scenarios and image restoration, respectively. To overcome the challenges posed by open-set scenarios, TDA introduces a degradation-agnostic adapter during the test phase (\textit{i.e.}, denoising process of PDM) for adapting PDM to the unknown and unforeseen degradations. This adapter is implemented by a simple neural network $\phi$, and optimized once after each denoising step towards aligning the domain of the $\hat{x}_0$ to that of the $y$. Through the adapter, $\hat{x}_0$ could be translated to the corresponding degraded one $\phi(\hat{x}_0)$, which could be further supervised towards the $y$, and in turn produce $\mathcal{G}$ to update the $\hat{x}_0$ towards the desired $x_0$. AIR is elaborated to conduct this supervised process for image restoration. Specifically, AIR is inspired by the observation that a PDM exhibits an intriguing temporal dynamic during the denoising process, \textit{i.e.}, $\hat{x}_0$ are generated from the unrecognizable contents to the perceptually rich contents, and finally the imperceptible details. Therefore, AIR dynamically adjusts the supervision strategies by gradually shifting the focuses from high- to low-level contents during the denoising process.

The framework of TAO is illustrated in Fig.\ref{fig:arch}, where a PDM is fixed to progressively denoising the pure noise $x_T$ to the clean image $x_0$. For the test degraded image $y$, TAO first samples two maps of random noise at $t=T$, and thus two images will be generated after each denoising step. We refer to the image updated towards the clean image of the $y$ as the guided image $x^g_t$, and the other one as the auxiliary image $x^a_t$. We introduce the $x^a_t$ to prevent TDA from falling into trivial solutions, and assist AIR to guide the $x^g_t$ towards the desired clean image $x^g_0$. At each step $t$, the images $x^g_t$, $x^a_t$, and $y$ are first fed into TDA for optimizing the adapter $\phi$ through the objectives for domain alignment, and then sent into AIR for updating $x^g_t$ towards $x^g_0$, which best matches $y$ undergoing $\phi$, through the conditional guidance in Eq.(\ref{eq:gui}).

\subsection{Test-time Degradation Adapter}

The essence of OIR lies in addressing the unidentified distribution shifts between the test and training data, while TTA recently emerges as an effective methodology to address this inherent disparities. With the motivations, TDA is presented for adapting PDM to the downstream shifted distribution of degradations during the test phase.

TDA employs a four-layer convolution neural network as the adapter $\phi$, which is optimized for aligning the domain of the generated images $\hat{x}^g_0$, $\hat{x}^a_0$ to that of the test degraded image $y$. Note that $\hat{x}^g_0$ and $\hat{x}^a_0$ are the estimated outputs from $x^g_t$ and $x^a_t$ in terms of Eq.(\ref{eq:deg}), respectively. Specifically, we introduce adversarial training which has shown to be effective in aligning the source and target domains, and the translated samples usually capture rich visual characteristics inherent to the target domain. Therefore, TDA aligns $\phi([\hat{x}^g_0, \hat{x}^a_0])$ to $y$ which involves significant visual degradations through
\begin{equation}
   \mathcal{L}_{adv} = \log(1-\mathcal{D}(\phi([\hat{x}^g_0, \hat{x}^a_0])),
\end{equation}
where $[\cdot]$ denotes the concatenation along the dimension of batch size, and $\mathcal{D}$ is a discriminator optimized through
\begin{equation}
   \mathcal{L}_{dis} = -\log \mathcal{D}([y, y]) - \log(1-\mathcal{D}(\phi([\hat{x}^g_0, \hat{x}^a_0])).
\end{equation}
Meanwhile, TDA adopts a reconstruction loss to maintain the content consistency during the domain alignment, \textit{i.e.},
\begin{equation}
    \mathcal{L}_{rec} = \parallel [\hat{x}^g_0, \hat{x}^a_0] - \phi([\hat{x}^g_0, \hat{x}^a_0]) \parallel_2^2.
\end{equation}
In addition, the perceptual loss is introduced to remain the semantic consistency during the domain alignment, \textit{i.e.},
\begin{equation}
    \mathcal{L}_{pec} = \parallel \mathcal{V}([\hat{x}^g_0, \hat{x}^a_0]) - \mathcal{V}(\phi([\hat{x}^g_0, \hat{x}^a_0])) \parallel_2^2,
\end{equation}
where $\mathcal{V}(\cdot)$ denotes the feature maps extracted from the pretrained VGG network. Overall, the objective for optimizing the adapter $\phi$ is
\begin{equation}
    \mathcal{L}_\phi = \lambda_1 * \mathcal{L}_{rec} + \lambda_2 * \mathcal{L}_{pec} + \lambda_3 * \mathcal{L}_{adv},
\end{equation}
where $\lambda_1$, $\lambda_2$, and $\lambda_3$ are the loss weights. By iterative optimization during the denoising process, $\phi$ gradually translates $\hat{x}^g_0$, $\hat{x}^a_0$ into the domain of $y$. Note that $\hat{x}^a_0$ is introduced to prevent $\phi$ from the trivial solution of identity mapping.

\subsection{Adapter-guided Image Restoration}

With the adapter $\phi$, $\hat{x}^g_0$ could be translated into the degraded domain of $y$ through $\phi(\hat{x}^g_0)$, which could be further supervised towards the $y$, and in turn produce the $\mathcal{G}$ to update the generated images $\hat{x}^g_0$ to the desired clean output $x^g_0$. AIR is devised to perform this supervised updating process.

Existing studies~\cite{choi2022perception} have show that the denoising process exhibits an intriguing temporal dynamic, \textit{i.e.}, $\hat{x}^g_0$ is generated from the unrecognizable contents to the perceptually rich contents, and finally the imperceptible details. In other words, the contents to be restored are varying as the denoising steps. To take full advantages of PDM, AIR empirically divides the denoising steps into three stages, and guides the restoration through different strategies.

In the first stage where $t \sim T$, the generated images have a lower signal-to-noise ratio, wherein the contents are unrecognizable and only contain some global information such as colors, layouts distributed throughout image. Therefore, AIR applies mse loss on the spatial pixels and color channels to learn global contents, \textit{i.e.},
\begin{equation}
    \mathcal{G} = \gamma_1 * \triangledown_{\hat{x}_0} \parallel y - \phi(\hat{x}_0) \parallel_2^2,
\end{equation}
where $\gamma_1$ is the loss weight and $\hat{x}_0 \in \{\hat{x}^g_0, \hat{x}^a_0\}$. Note that AIR also use this guidance on the auxiliary image $x^a_0$ so that it involves the similar global properties as $y$, since it is beneficial to the optimization of adapter on most degradations. In addition, a linear warmup is employed on $\gamma_1$ to gradually involve this guidance.

In the second stage where $t \sim T/2$, PDM prefers to generate the perceptually rich contents, and thus AIR focuses more on restoring the perceptual contents from $y$. To this end, the perceptual loss is adopted, \textit{i.e.},
\begin{equation}
    \mathcal{G}_{pec} = \triangledown_{\hat{x}^g_0} \parallel \mathcal{V}(y) - \mathcal{V}(\phi(\hat{x}^g_0)) \parallel_2^2.
\end{equation}
Meanwhile, an adversarial loss is also introduced to boost the perceptual quality, \textit{i.e.},
\begin{equation}
    \begin{aligned}
        \mathcal{G}_{adv} = & \triangledown_{\hat{x}^g_0} \log(1-\mathcal{D}(\hat{x}^g_0 - y)) \\
        \mathcal{L}_{dis} = - &\log(1-\mathcal{D}(\hat{x}^g_0-y))\\
        - &\log\mathcal{D}([\hat{x}^g_0,\hat{x}^a_0]-\phi([\hat{x}^g_0,\hat{x}^a_0])).
    \end{aligned}
\end{equation}
With this loss, $\hat{x}^g_0$ will be further updated toward the clean image whose degraded version matches $y$, by aligning the domain of content loss caused by degradations. Besides, the mse loss is also employed to remain the global contents, \textit{i.e.},
\begin{equation}
    \mathcal{G}_{mse} = \triangledown_{\hat{x}^g_0} \parallel y - \phi(\hat{x}^g_0) \parallel_2^2.
\end{equation}
Overall, the guidance loss for the second stage is
\begin{equation}
    \mathcal{G} = \gamma_2 * \mathcal{G}_{mse} + \gamma_3 * \mathcal{G}_{pec} + \gamma_4 * \mathcal{G}_{adv},
\end{equation}
where $\gamma_{2-4}$ are the loss weights.

In the third stage where $t \sim 0$, the perceptually rich contents are already presented in $\hat{x}^g_0$. AIR is only required to recover the imperceptible details. In other words, $\hat{x}^g_0$ is very close to the natural image, and thus some extra image priors could be introduced to regulate the final output. In consideration of generality, we only introduce a total variation loss, \textit{i.e.}, $\mathcal{G} = \gamma_5 * TVLoss(\hat{x}^g_0)$, where $\gamma_5$ is the loss weight. Note that although the third stage mainly affects the imperceptible details, the pixel values still fluctuate and thus the guidance loss in the second stage is used for a better numerical fidelity.

\section{Experiments}
\label{sec:experiments}

In this section, we first introduce the experimental settings, and then show quantitative and qualitative results on multiple degradations. Finally, we perform analysis experiments including ablation studies and result visualizations.

\begin{table*}[!htbp]
\centering
\caption{Quantitative results of image dehazing on HSTS. Our method outperforms zero-shot methods, and obtains comparable even better results than classic methods through supervised learning. All the compared methods are specifically designed for image dehazing.}
\vspace{0.2cm}
\begin{tabular}{r|cccc|cccccc|c}
\toprule
\multirow{2}{*}{Metrics} & \multicolumn{4}{c|}{Classic Learning Methods} & \multicolumn{6}{c}{Zero-shot Methods} & \multicolumn{1}{|c}{OIR} \\
\cmidrule{2-12}
& DEN & MSN & AOD & CAP & DCP & BCCR & GRM & NLD & DDIP & YOLY & Ours \\
\midrule
PSNR~$\uparrow$ & 23.96 & 18.40 & 19.15 & 21.69 & 18.42 & 15.75 & 17.41 & 18.18 & 18.22 & 21.24 & 22.29 \\ 
SSIM~$\uparrow$ & 0.902 & 0.826 & 0.860 & 0.868 & 0.854 & 0.783 & 0.833 & 0.814 & 0.832 & 0.835 & 0.877 \\ 
\bottomrule
\end{tabular}
\label{tab:dh}
\end{table*}

\begin{figure*}[!htbp]
    \centering
    \includegraphics[width=0.99\textwidth]{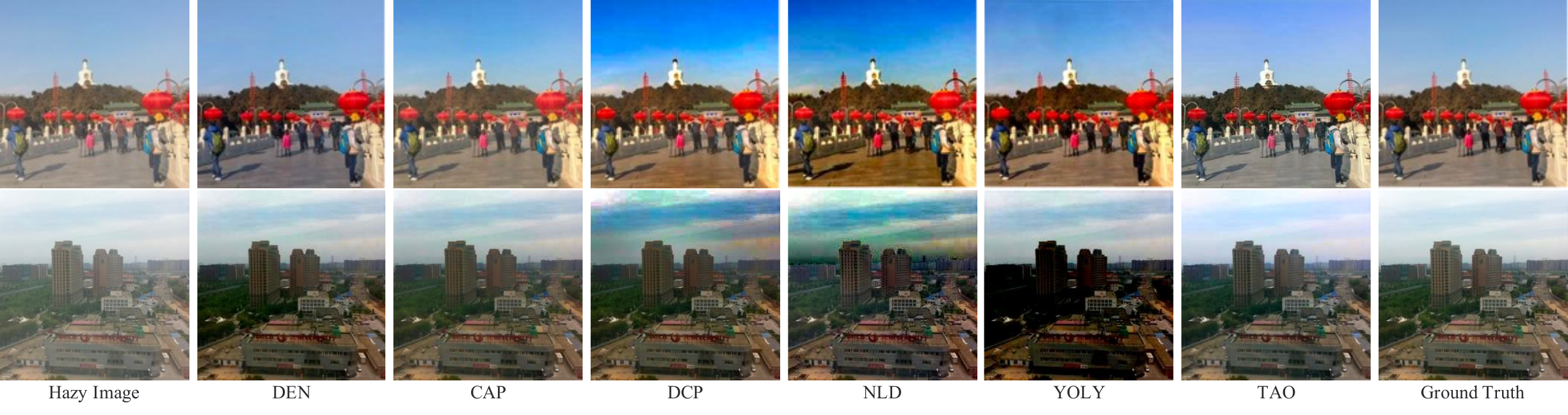}
    \vspace{-0.2cm}
    \caption{Qualitative results on image dehazing, from which one could observe that existing methods excessively dehazing resulting in darkening and/or artifacting of the images. In contrast, our method obtains clearer results which are closer to the natural ground truths.}
    \label{fig:dehazing}
\end{figure*}

\subsection{Experimental Settings}

To evaluate the effectiveness of TAO on addressing OIR, we conduct experiments on multiple tasks of image restoration with exactly the same settings, except for the loss weights and guidance scale ($\lambda_{1-3}$, $\gamma_{1-5}$, $s$), which are finetuned for different types of degradation to obtain the best results. In experiments, we employ an unconditional image diffusion model~\cite{guidiff} pretrained on ImageNet~\cite{imagenet} as PDM, and sets the timestep as $T=1000$, which is further divided into three stages in a heuristic way, \textit{i.e.}, the first stage 999-700, the second stage 700-50, and the third stage 50-0. Since PDM is pre-trained on large-scale natural images for high-quality image generation, any degradation in the test image could be regraded as the unforeseen. Both the adapter and discriminators are four-layer convolutional networks, and optimized once at each denoising step through Adam optimizer with default learning rate of 1e-3. First nine layers of pretrained VGG-16~\cite{vgg16} network are employed to extract the semantic feature maps for calculating the perceptual loss. For evaluations, considering the specificity of our methods, \textit{i.e.}, adaptation on single test sample, we mainly compare it with those task-specific zero-shot methods, and some classic learning-based methods. To access their performance, PSNR and SSIM metrics are employed, and all experiments are conducted through PyTorch framework on Ubuntu20.04 with GeForce RTX 3090 GPUs.

\subsection{Experimental Results}

Here, we evaluate our method on the tasks of image dehazing, low-light image enhancement, and image denoising.

\textbf{Image Dehazing} aims to eliminate the haze and boost the visibility of hazy image, which has spawned many methods in the past decades. In experiments, we introduce the HSTS dataset from RESIDE~\cite{reside} for evaluations. To be specific, RESIDE is a large scale haze image dataset, and HSTS is one of the test subsets, which contains 10 synthetic and 10 real-world hazy images. Before the evaluation, we first center crop the images along the shorter edges, and then resize them to match the image size of the PDM.

For a comprehensive comparison, we compare TAO with 10 representative methods which are specifically designed for image dehazing, and could be roughly divided into two groups, \textit{i.e.}, classic learning-based and zero-shot methods. Specifically, classic learning-based methods are DehazeNet (DEN)~\cite{dehazenet}, MSCNN (MSN) \cite{mscnn}, AOD-Net (AOD)~\cite{aod}, and CAP~\cite{cap}. Zero-shot methods are DCP~\cite{dcp}, BCCR~\cite{bccr}, GRM~\cite{grm}, NLD~\cite{nld}, DDIP~\cite{ddip} and YOLY~\cite{yoly}.

The quantitative results are presented in Tab.\ref{tab:dh}, from which one could see that our method achieves comparable even better results than those task-specific methods. To be specific, our method obtains the best performance in zero-shot methods, and outperforms YOLY and DCP with 1.05/0.042 and 3.87/0.023 according to PSNR/SSIM metrics, respectively. Meanwhile, even compared with the classic learning-based methods, our method still outperforms most of them and achieves comparable results with the others. For example, our method obtains 3.89/0.051, 3.14/0.017, and 0.60/0.009 higher PSNR/SSIM values than MSN, AOD and CAP, respectively. Although our method is lower than DEN, the performance gaps are not huge, \textit{e.g.}, 1.67/0.025 lower than DEN in terms of PSNR/SSIM metrics. In addition, we also show and analyze the qualitative results in Fig.\ref{fig:dehazing}, from which one could observe that our method achieves better fidelity and realness than the compared methods.

\begin{table*}[!htbp]
\centering
\caption{Quantitative results of low-light image enhancement on LOL dataset. Our method obtains the best SSIM values and the second PSNR values in the zero-shot methods, and only the classic learning-based MBLLEN outperforms our method on both the two metrics.}
\vspace{0.2cm}
\begin{tabular}{r|cccc|ccccc|c}
\toprule
\multirow{2}{*}{Metrics} & \multicolumn{4}{c|}{Classic Learning Methods} & \multicolumn{5}{c|}{Zero-shot Methods} & OIR \\
\cmidrule{2-11}
& LNet & RNet & MBLLEN & EGAN & ExCNet & ZDCE & ZDCE+ & RRDN & GDP & Ours \\
\midrule
PSNR~$\uparrow$ & 13.39 & 17.87 & 18.03 & 14.85 & 17.15 & 15.33 & 15.71 & 11.24 & 15.75 & 17.10 \\
SSIM~$\uparrow$ & 0.623 & 0.699 & 0.809 & 0.815 & 0.720 & 0.763 & 0.765 & 0.543 & 0.665 & 0.804 \\
\bottomrule
\end{tabular}
\label{tab:ll}
\end{table*}

\begin{figure*}[!htbp]
    \centering
    \includegraphics[width=0.93\textwidth]{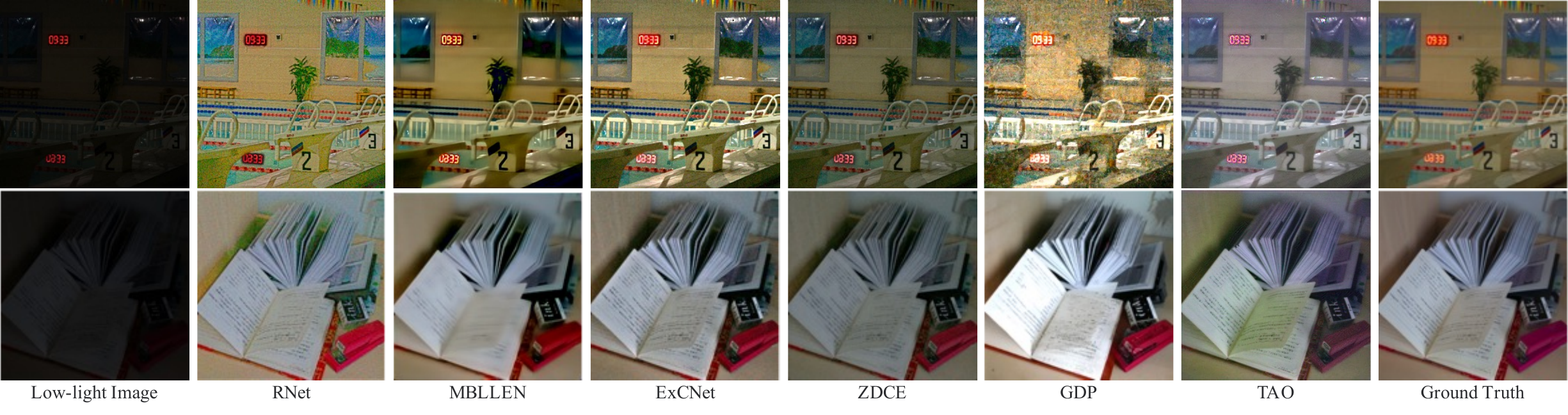}
    \vspace{-0.2cm}
    \caption{Qualitative results on low-light image enhancement, from which one could see that our results are not as smooth as MBLLEN nor as dark as ZDEC. Although there are slight color biases from ground truths, our method achieves a rational lighting of the dark images.}
    \label{fig:lowlight}
\end{figure*}

\textbf{Low-light Image Enhancement} aims at improving the perception of the image captured in the environment with poor illumination. To evaluate our method on this task, we introduce the test subset from LOL~\cite{wei2018deep} dataset. LOL includes 485 training and 15 test image pairs of low- and normal-light, which contain noises produced during the photo capture process. Since the images have a resolution of 400$\times$600, we first center crop them along the shorter edges, and then resize them to be applicable to the PDM.

For a comprehensive comparison, we compare our method with nine low-light image enhancement methods, which are divided into two categories, \textit{i.e.}, classical learning-based and zero-shot methods. To be specific, the classical learning-based methods are LightenNet (LNet)~\cite{lnet}, Retinex-Net (RNet)~\cite{rnet}, MBLLEN~\cite{mbllen}, and EnlightenGAN (EGAN)~\cite{egan}. The zero-shot methods are ExCNet~\cite{excnet}, Zero-DCE (ZDEC)~\cite{zdec}, Zero-DCE++ (ZDEC+)~\cite{zdec+}, RRDNet (RRDN)~\cite{rrdn}, and GDP~\cite{gdp}. The above methods except GDP are specially designed for low-light image enhancement, while our method is devised for more general purpose of OIR, and the only difference for different tasks lies in the loss weights and the guidance scale.

The quantitative results are shown in Tab.\ref{tab:ll}. From the Table, one could see that our method achieves comparable even better results than those specific-designed methods. In the zero-shot methods, our method achieves the best and the second performance in terms of SSIM and PSNR metrics, respectively. In PSNR, our method outperforms the other methods except ExCNet with the margins of 1.35-5.86. Although ExCNet achieves a slightly better PSNR value of 0.05, our method achieves the best SSIM values and exceeds it with a margin of 0.084. Besides, even compared with the classic learning-based methods, our method is competitive and obtains better results than most of them. For instance, our method obtains 3.71/0.181 higher PSNR/SSIM value than LNet, 2.25 higher PSNR value than EGAN, and 0.105 higher SSIM value than RNet. Only MBLLEN outperforms our method on both the PSNR and SSIM metrics with the slight margins of 0.93 and 0.005, respectively. In addition, we present and analyze the qualitative results in Fig.\ref{fig:lowlight}, which demonstrate our method enlightens the dark images reliably and reasonably.

\begin{table}[!htbp]
\centering
\caption{Quantitative results of image denoising on Kodak dataset. N.I. denotes the noise images. From the table, one could see that our method is comparable to even better than those zero-shot image denoising methods.}
\vspace{0.2cm}
\resizebox{\linewidth}{!}{\begin{tabular}{r|cccccc}
\toprule
Metrics & N.I. & BM3D & N2S & N2V & DIP & Ours \\
\hline
PSNR~$\uparrow$ & 26.35 & 28.63 & 28.71 & 27.17 & 27.58 & 28.49 \\
SSIM~$\uparrow$ & 0.757 & 0.795 & 0.864 & 0.846 & 0.808 & 0.830 \\
\bottomrule
\end{tabular}}
\label{tab:dn}
\end{table}

\begin{figure*}[!htbp]
    \centering
    \includegraphics[width=0.83\textwidth]{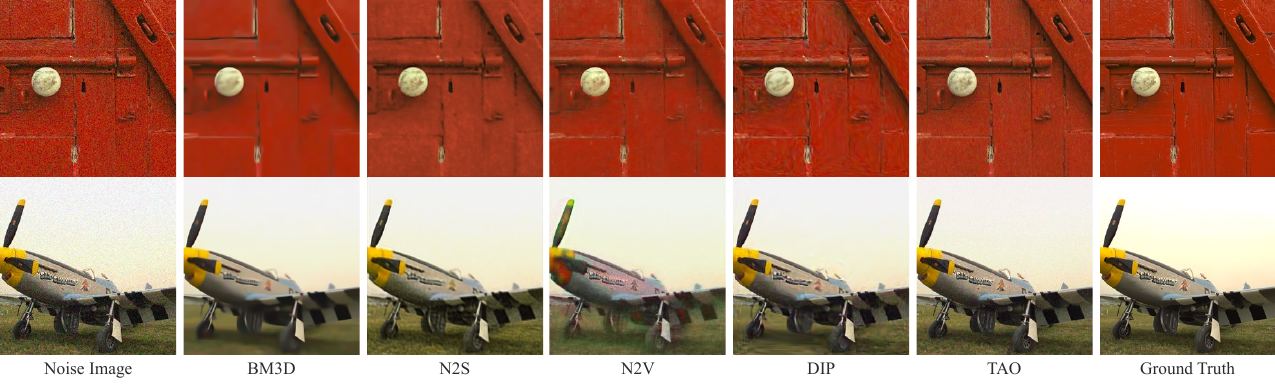}
    \vspace{-0.2cm}
    \caption{Qualitative results on image denoising, from which one could observe that existing methods excessively denoising resulting in smoothing and/or artifacting of the images. In contrast, our method obtains clearer and sharper results which are closer to ground truths.}
    \label{fig:denoising}
\end{figure*}

\begin{figure}[!htbp]
    \centering
    \includegraphics[width=0.95\linewidth]{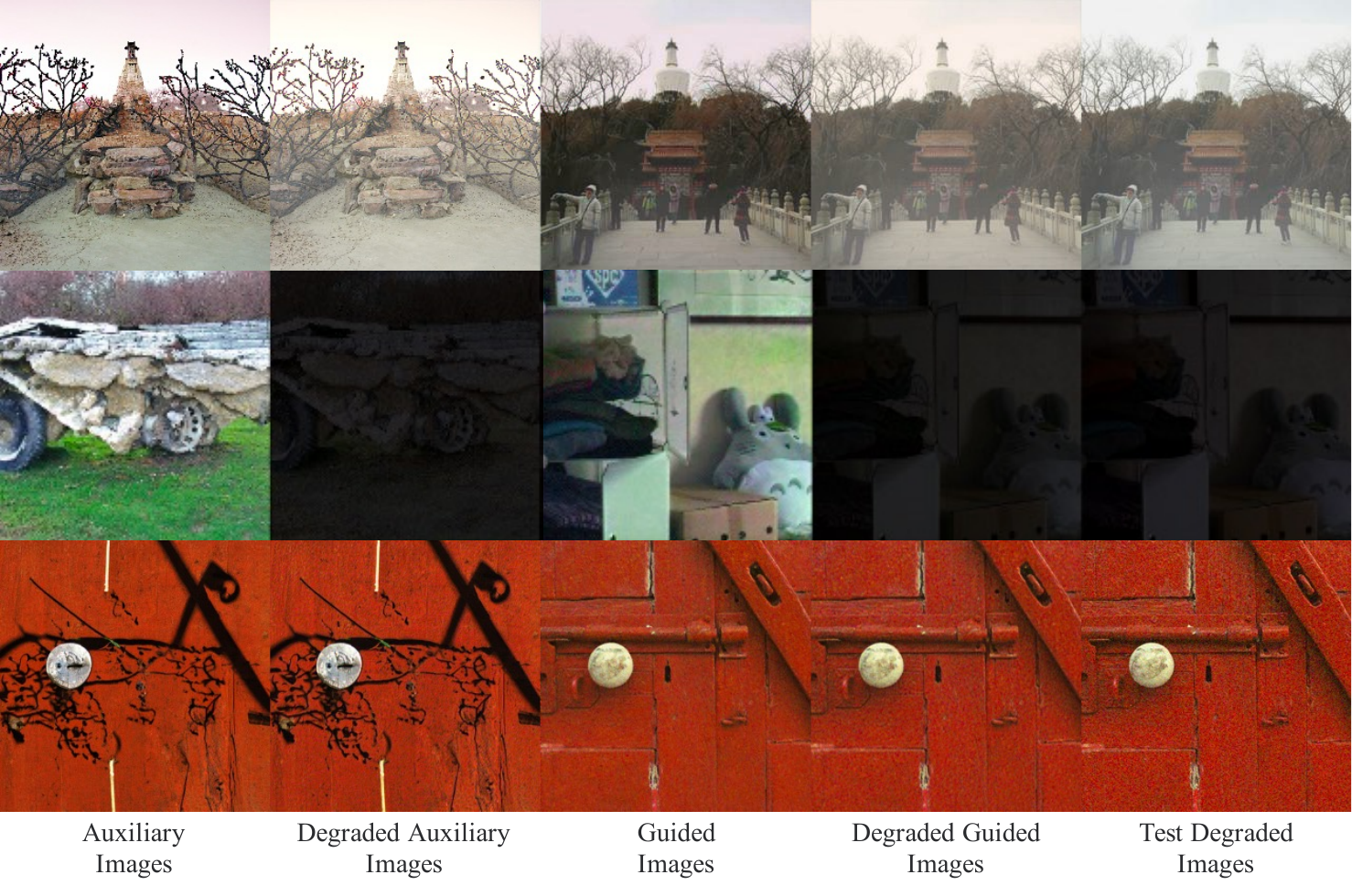}
    \vspace{-0.2cm}
    \caption{Visualizations from TDA wherein Auxiliary/Guided Images are translated to Degraded Auxiliary/Guided Images by TDA, whose domains align well to that of Test Degraded Images.}
    \label{fig:tda}
\end{figure}

\begin{table*}[!htbp]
\centering
\caption{Results on different strategies for domain alignment to unknown and unseen degradations. These strategies were implemented based on BasicSR~\cite{basicsr}. VGAN-S/B denote the Sigmoid Layer plus BCELoss and BCEWithLogitsLoss, respectively.}
\vspace{0.2cm}
\resizebox{0.83\textwidth}{!}{\begin{tabular}{c|ccccccc}
\toprule
Metrics & Haze Image & VGAN\_S & VGAN\_B & LSGAN & WGAN & WGAN\_Softplus & Hinge  \\
\hline
PSNR~$\uparrow$ & 14.74 & 22.29   & 22.06   & 20.76  & 13.61  & 21.47   & 20.44  \\
SSIM~$\uparrow$ & 0.770 & 0.877  & 0.870  & 0.873 & 0.619 & 0.864  & 0.860 \\
\bottomrule
\end{tabular}}
\label{tab:adv}
\end{table*}

\textbf{Image Denoising} aims to remove noises from an noise image. To evaluate our method, we introduce Kodak24 dataset which consists of 24 natural clean images, and is commonly used for testing image denoising methods. Similarly, we center crop and resize the image to match the size of the PDM, and obtain the noise images by adding the Gaussian noises with the noise level of $\sigma=30$ to the clean images.

We compare our method with four zero-shot image denoising methods, \textit{i.e.}, BM3D~\cite{bm3d}, N2S~\cite{n2s}, N2V~\cite{n2v}, and DIP~\cite{dip}. The above methods except for DIP are specifically designed for image denoising. The quantitative results are presented in Tab.\ref{tab:dn}. From the table, one could see that our method obtains comparable even better results than those task-specific methods. To be specific, our method obtains 0.035 and 0.022 higher SSIM value than BM3D and DIP, respectively. Meanwhile, our method also outperforms N2V and DIP in PSNR value with a large margin of 1.32 and 0.91, respectively. Although not achieving the best performance, our method enjoys the appealing capacity of addressing OIR through one model. In addition, we present the qualitative results in Fig.\ref{fig:denoising}, from which one could see that our method obtains clearer results with better fidelity to ground truths than the compared methods.

\subsection{Analysis Experiments}

In this section, we conduct analysis experiments w.r.t. our proposed TDA and AIR mainly on image dehazing.

\textbf{The effectiveness of TDA.} Since PDM cannot achieve image dehazing without TDA, we demonstrate its effectiveness by i) observing whether it aligns the domain of generated images to that of the degraded image after the optimization during the denoising process, and ii) comparing with different strategies of adversarial training for domain alignment. The visual results are shown in Fig.\ref{fig:tda}, from which one could observe that TDA exhibits a significant capability of domain alignment to the unknown and unforeseen degradations. For example, the haze in the top row is uneven distributed in the test degraded image, and the same effects exist in the degraded auxiliary/guided image. The degree of darkness in the degraded auxiliary/guided image is similar to that in the test degraded image in the second row. In addition, we conduct experiments on different strategies of adversarial training implemented by BasicSR~\cite{basicsr}, and present the results in Tab.~\ref{tab:adv}. From the table, one could see that, except for WGAN, all strategies achieve the significant performance gains over the haze images. Namely, our adapter generalizes well to different strategies for domain alignment to the unknown and unseen degradations.

\begin{figure}[!htbp]
    \centering
    \includegraphics[width=0.99\linewidth]{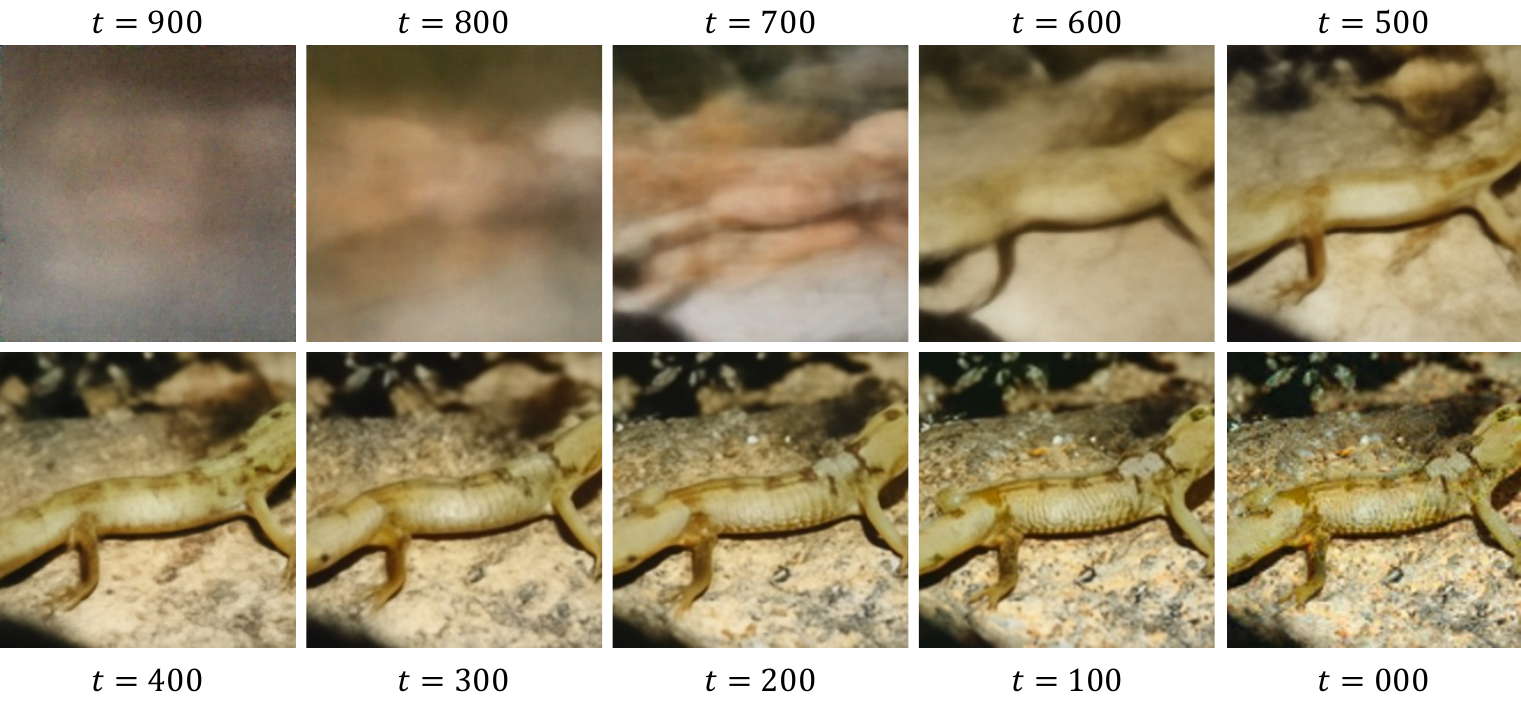}
    \vspace{-0.2cm}
    \caption{The temporal dynamics of PDM during the denoising process, which motivates our AIR to guide the image restoration through different strategies at different denoising steps.}
    \label{fig:motair}
\end{figure}

\begin{table}[!htbp]
\centering
\caption{Ablation studies on guidance strategies, where U.G. denotes using the same guidance loss of mean squared error throughout the denoising process. -F.S., -S.S., and -T.S. denote to remove the First Stage, Second Stage, and Third Stage, receptively. The results demonstrate the indispensable roles of the three stages for image dehazing.}
\vspace{0.2cm}
\resizebox{0.93\linewidth}{!}{\begin{tabular}{r|cccccc}
\toprule
Metrics & U.G. & -F.S. & -S.S. & -T.S. & AIR \\
\hline
PSNR~$\uparrow$ & 20.49 & 21.43 & 21.06 & 21.37 & 22.29 \\
SSIM~$\uparrow$ & 0.869 & 0.873 & 0.865 & 0.871 & 0.877 \\
\bottomrule
\end{tabular}}
\label{tab:absair}
\end{table}

\textbf{The effectiveness of AIR.} Here, we first demonstrate the rationality of AIR in Fig.\ref{fig:motair} which shows a typical denoising process of PDM. From the figure, one could see the significant temporal dynamics, \textit{i.e.}, the images are generated from the unrecognizable contents to the perceptually rich contents, and finally the imperceptible details. The dynamics also imply the varying of the contents should be focused and restored in the test degraded image, and thus applying different guidance strategies could facilitate image restoration. We present the ablation results in Tab.~\ref{tab:absair}, which show the effectiveness of the three stages on facilitating image dehazing. In addition, we also conduct analysis experiments on stage divisions as shown in Tab.~\ref{tab:absstg}, from which one could see that both early and late dividing steps result in the inferior performances. In other words, a proper stage division consistent with the temporal dynamics could improve the performance of image restoration.

\begin{table}[!htbp]
\centering
\caption{Analysis experiments on divisions of the three stages. The left is the dividing step between the First and Second stages, and the right is the dividing step between the Second and Third stages. The results demonstrate a proper stage division could significantly improve the restoration.}
\vspace{0.2cm}
\resizebox{0.99\linewidth}{!}{\begin{tabular}{c|ccc|ccc}
\toprule
Metrics & 800   & 700   & 600   & 70    & 50    & 30    \\
\hline
PSNR~$\uparrow$ & 21.89 & 22.29 & 21.11 & 21.76 & 22.29 & 22.01 \\
SSIM~$\uparrow$ & 0.861 & 0.877 & 0.868 & 0.876 & 0.877 & 0.872 \\
\bottomrule
\end{tabular}}
\label{tab:absstg}
\end{table}

\section{Conclusion}
\label{sec:conclusion}

This paper explicitly explores the challenges posed by open-set scenarios, and formally defines the problem of open-set image restoration. To solve this problem, this work reveals its essence from the perspective of distribution shifts, and discovers the methodology of test-time adaptation is adept at addressing this inherent disparities. Motivated by this, we presented a test-time degradation adaptation framework for open-set image restoration, which is ingenious in the following ways. First, it considers a pre-trained image diffusion model as the general pretraining model for solving various tasks of image restoration. Second, it introduces an adapter optimized during the test phase for adapting the pre-trained model to the unknown and unseen test degradations. Third, it dynamically adjusts the guidance strategies following the denoising process to obtain better restoration results. Through experiments on multiple degradations, we demonstrate the effectiveness of our designs.

\section*{Impact Statement}

This paper considers a novel problem in Image Restoration that restores the clean images from the degraded ones under an open-set scenario. Although there are potential societal consequences of this work, none which we feel serious and must be specifically highlighted here. For example, the most serious cases may be the recovery of images that have been intentionally damaged by someone else, or do not match the facts affecting the decisions made on that.

\section*{Acknowledgment}

The authors would like to thank the anonymous reviewers and area chair for their valuable comments and constructive suggestions to improve the quality of this paper. This work was supported in part by NSFC under Grant 62176171, U21B2040; in part by the Fundamental Research Funds for the Central Universities under Grant CJ202303; and in part by Sichuan Science and Technology Planning Project under Grant 24NSFTD0130.


\bibliography{example_paper}
\bibliographystyle{icml2024}

%
%

\end{document}